\documentclass{article} % For LaTeX2e
\usepackage[preprint]{colm2026_conference}

\usepackage{microtype}
\usepackage{hyperref}
\usepackage{url}
\usepackage{booktabs}
\usepackage{amsmath}
\usepackage{graphicx}
\usepackage{natbib}
\usepackage{lineno}

\definecolor{darkblue}{rgb}{0, 0, 0.5}
\hypersetup{colorlinks=true, citecolor=darkblue, linkcolor=darkblue, urlcolor=darkblue}

\title{Semantic Structure of Feature Space in \\Large Language Models}

\author{Austin C. Kozlowski\\
Knowledge Lab\\
University of Chicago\\
\And
Andrei Boutyline \\
Department of Sociology \\
University of Michigan \\
}

\begin{document}

\ifcolmsubmission
\linenumbers
\fi

\maketitle

\begin{abstract}
 We show that the geometric relations between semantic features in large language models' hidden states closely mirror human psychological associations. We construct feature vectors corresponding to 360 words and project them on 32 semantic axes (e.g. \textit{beautiful-ugly}, \textit{soft-hard}), and find that these projections correlate highly with human ratings of those words on the respective semantic scales. Second, we find that the cosine similarities between the semantic axes themselves are highly predictive of the correlations between these scales in the survey. Third, we show that substantial variance across the 32 semantic axes lies on a low-dimensional subspace, reproducing patterns typical of human semantic associations. Finally, we demonstrate that steering a word on one semantic axis causes spillover effects on the model’s rating of that word on other semantic scales proportionate to the cosine similarity between those semantic axes. These findings suggest that features should be understood not only in isolation but through their geometric relations and the meaningful subspaces they form.\end{abstract}

\section{Introduction}

A growing literature shows that many features, such as deception, refusal, and sycophancy, are encoded as linear directions in large language models’ (LLMs) internal representations \citep{arditi2024refusal,bricken2023monosemanticity,park2024geometry}. This line of work has produced useful tools for auditing and controlling LLMs by isolating features of interest and then monitoring, amplifying, or suppressing them. Linear features thus at times serve as a vast collection of knobs that can be individually tweaked to modify LLM outputs \citep{durmus2024steering,lieberum2024gemma,templeton2024scaling}. 

However, considering features in isolation ignores a key characteristic of distributed representations – that many features are not rendered orthogonally, and that proximities and directions in the representation space encode meaningful relations between features. In the case of modeling language, this was first evident with early studies of word embedding models like word2vec \citep{mikolov2013efficient,mikolov2013distributed}. Cosine similarities between word vectors captured semantic and syntactic similarities, and directions in the space corresponded to general features like gender or nationality \citep{bolukbasi2016man,caliskan2017semantics,garg2018word,pennington2014glove}. Subsequent work combined these insights to show that, these general linear features, like masculine-feminine, good-bad, or rich-poor, are themselves meaningfully non-orthogonal, their cosine similarities reflecting cultural associations such as gender bias or markers of social class \citep{boutyline2023school,kozlowski2019geometry}. Moreover, recent evidence shows that the first-layer token embeddings of LLMs preserve meaningful geometric feature structures much like classic word embeddings \citep{kozlowski2025semantic}. However, there remains a paucity of research investigating whether these semantic structures persist in LLMs' middle layers, where features are most commonly monitored and steered \citep{skean2025layer}. 

Classic studies from psychology provide more reason to expect that important semantic features may be entangled in LLMs' internal representations. Studies in the ``semantic differential'' tradition had respondents rate large sets of words on a variety of semantic scales, such as masculine-feminine, soft-hard, and fast-slow \citep{jenkins1958atlas,osgood1957measurement,osgood1964semantic}, and consistently found these data reduced effectively to a 3-dimensional solution, with the most explanatory components corresponding to Evaluation (good-bad), Potency (strong-weak), and Activity (active-passive) \citep{combs2025deviations, heise2010surveying}. This core finding suggests that the apparent diversity of features that humans use to make sense of the world can be largely captured by a relatively low-dimensional subspace.

Motivated by these lines of inquiry, we examine the geometric relations between semantic features in LLMs’ internal representations, then compare these relations to human psychological associations measured with a survey. We find that the geometry of the LLM feature space closely reproduces human associations in several respects. First, the projection of features along constructed ``semantic axes'' correlates highly with average human ratings of the corresponding words on respective semantic scales. Second, we show that the cosine similarities between the semantic axes themselves are substantial, and that these cosine similarities are highly predictive of the correlation between the corresponding scales in the semantic survey. Third, we apply principal components analysis (PCA) to these semantic axes, finding that a substantial share of their variance is explained by the first three components and that the subspace defined by the first three components is closely aligned with that derived from the survey data.

Finally, we find that the proximities between semantic axes have practical implications for feature steering \citep{durmus2024steering,panickssery2023steering}. Steering along one semantic axis induces spillover effects on how the model rates words on other semantic axes proportional to the cosine similarity between those axes. This implies that the cosine similarities between distinct semantic features are substantial enough to influence model behavior. These findings together suggest that the geometry of feature space is humanly meaningful, and that features need not be only considered in isolation; considering angles between features or the subspaces that preserve multiple features could offer valuable insight into a feature’s meaning and operation within an LLM.

\section{Data \& Methods}

Our analyses draw upon two forms of data: human responses to a survey of semantic associations, and values derived from LLM activations and outputs.

\subsection{Survey of Semantic Associations}

To assess human associations between semantic features, we use data from a recent survey \citep{boutyline2025forging}. The survey was fielded on the online platform Prolific to a quota sample of 1,750 respondents approximately representative of the U.S. population along major demographic categories. Respondents each rated a subset of all word/scale pairs, and each word was ultimately rated on each scale by an average of 24 respondents. The survey is a modern replication of a classic semantic differential survey and reused the same set of 360 words and 20 scales. These scales were originally selected to ``exhaust the semantic space most thoroughly'' (p. 689) and the words were selected based on their high variance across several other psychological inventories \citep{jenkins1958atlas}. The modern replication added 16 new scales to capture more social attributes (e.g. ``free-unfree'', ``equal-unequal''). We exclude four scales that were insufficiently distinct from other scales analyzed (e.g. human-machine with natural-artificial; trustworthy-untrustworthy with true-untrue). Our analytic sample therefore comprises 360 words rated on 32 semantic scales.

\subsection{LLM Features}
We run our experiments on small and large models from the Llama family (Llama 3.2 3B and Llama 3.1 70B) \citep{grattafiori2024llama}. The same analyses applied to Qwen3 1.7B and Qwen3 32B produce similar results and are included in the Appendix \citep{yang2025qwen3}.

To facilitate direct comparison to human responses, we create a dataset of semantic associations derived from LLM activation patterns that matches the structure of the survey data. We constructed feature vectors for each of the 360 words rated in the survey and for 32 ``semantic axes'' corresponding to the survey’s semantic scales. To construct feature vectors for the 360 words, we followed a simple protocol based on the method described and proven effective in \cite{lindsey2026emergent}. For each word, we passed the following four prompts through the LLMs and collected their activations:

\begin{quote}
\textbf{USER:} Tell me about \{\textit{word\(_i\)}\}.

\textbf{USER:} Tell me your associations with the word \{\textit{word\(_i\)}\}.

\textbf{USER:} What do you think of when you hear the word \{\textit{word\(_i\)}\}?

\textbf{USER:} Define the word \{\textit{word\(_i\)}\}. \textbf{ASSISTANT:} The word \{\textit{word\(_i\)}\} means
\end{quote}

Our preliminary investigations found that semantic structure emerges clearly at different depths for small and large models. In accordance with these findings, we analyze activations from layer 8 (30\% depth) in the 3B Llama model and layer 40 (50\% depth) in the 70B model. To construct a single feature vector, we calculate the mean activation across all tokens for each prompt. After the feature vector is calculated for each of the four prompts, we take the mean of these four vectors to get a single robust feature vector for each of the 360 words. Robustness tests demonstrated that the four prompts all perform similarly and that results are consistent if activations are collected from the prompt’s last token activation rather than through mean pooling across tokens.

To construct the semantic axes, we use a contrastive pair design \citep{panickssery2023steering,tigges2023linear}. We compile sets of 10 antonym pairs designating each semantic axis (e.g. good – bad, great – terrible, etc.). For each of these individual antonym words, we construct a feature vector by the same procedure described above for the 360 words. We then calculate the differences between each of the ten pairs of antonym feature vectors and take the mean of these ten differences. 

\begin{equation}
\mathbf{sem\_axis}_f = \frac{1}{10} \sum_{j=1}^{10} \left( \mathbf{pos\_antonym}_{j} - \mathbf{neg\_antonym}_{j} \right)
\label{eq:sem_axis}
\end{equation}

\subsection{Measuring Semantic Structure}
\begin{figure}[h!]
    \centering
    \includegraphics[width=\textwidth]{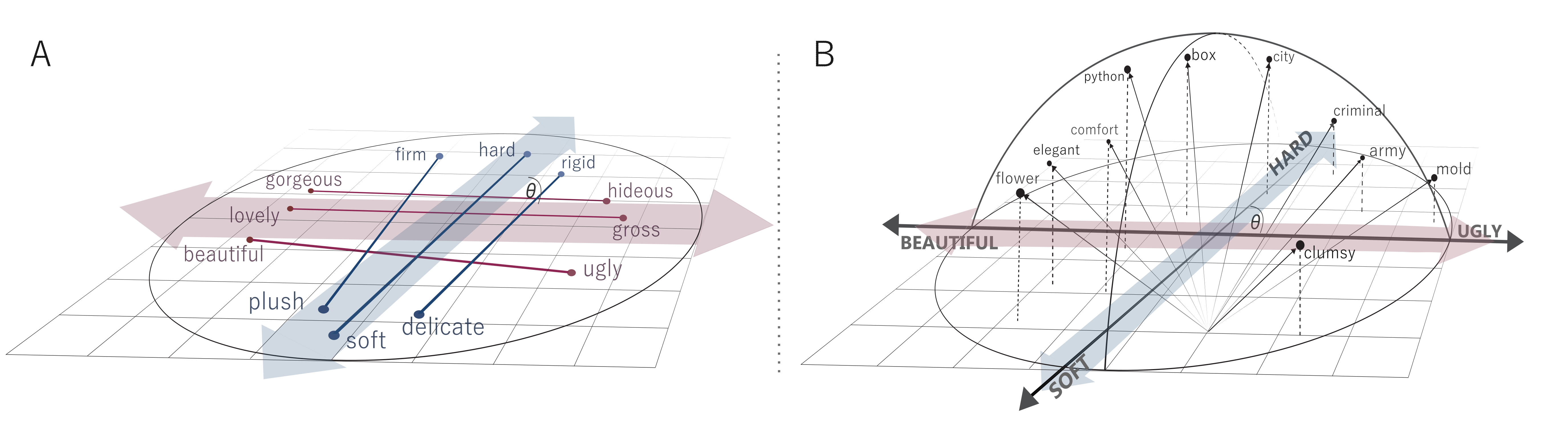}
    \caption{Conceptual diagram representing (A) the construction of semantic axes with antonym pairs and (B) the projection of word feature vectors on semantic axes.}
    \label{fig:space_schematic}
\end{figure}

We take two approaches to measuring semantic structure in LLM feature space. First we examine how features corresponding to individual words like "army" or "city" are located along semantic feature vectors, and test if their projections on semantic axes correlate with human ratings of those words on semantic scales. Second, we measure the cosine similarity between the semantic axes themselves, and test if the geometric alignment of features in the activation space predicts the correlation between the respective semantic scales in the human survey data. The benefit of this test is that it goes beyond assessing whether there is a correlation between the data points on these axes and tests whether the axes themselves are meaningfully non-orthogonal (see Figure \ref{fig:space_schematic}).

We next use PCA on both the survey dataset and the LLM feature vectors. Prior psychological studies confirm that human semantic ratings reduce effectively to a 3-dimensional subspace \citep{heise2010surveying,osgood1957measurement}. After replicating this on our survey dataset, we perform PCA on our LLM feature vectors. In addition to applying PCA on the projections of the 360 words on the 32 semantic axes, we also apply it to the semantic axis vectors themselves to examine the subspace defined by the raw vectors rather than the projection of a specific set of words upon them. We then use canonical correlation analysis (CCA) to determine how well the 3-dimensional subspace defined by the LLM's semantic axes aligns with the subspace derived from the survey data.

\subsection{Steering Interventions}
Lastly, we conduct steering interventions to test if the geometric alignment of features has practical implications for model behavior. To measure model behavior, we prompt the model to provide an association for each of the 360 words along each of the semantic scales. The prompt takes the following form:

\begin{quote}
\textbf{USER:} Do you associate '\{\textit{word}\}' more with \{\textit{pos\_antonym}\} or \{\textit{neg\_antonym}\}? Just print your assessment with no formatting.

\textbf{ASSISTANT:} I find \{\textit{word}\} to be more
\end{quote}

We iterate over all 360 words for each of the 32 antonym pairs used in the survey. For each prompt, we extract the logits associated with \textit{pos\_antonym} and \textit{neg\_antonym} (for example, "beautiful" and "ugly") output by the model to predict the next token. We then apply a softmax transformation over those logits to transform them into probabilities.

\begin{figure}[h!]
    \centering
    \includegraphics[width=\textwidth]{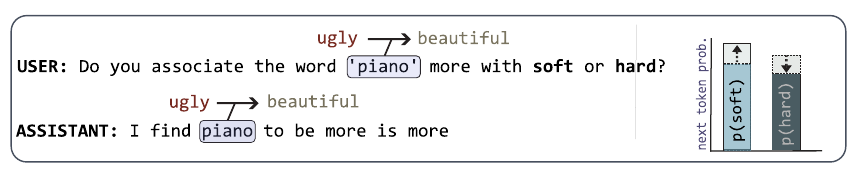}
    \caption{Conceptual diagram representing the format of prompts and the application of steering interventions used to test for spillover effects between semantic axes. As "piano" is steered toward \textit{beautiful} and away from \textit{ugly}, the model is likely to describe it as \textit{soft}.}
    \label{fig:prompt_schematic}
\end{figure}

After collecting baseline probabilities for each word’s associations along each of the semantic scales, we intervene on the forward pass with each semantic axis feature vector. We implement a forward hook on the residual stream at a single layer (at 30\% depth for the 3B model and 50\% depth for 70B, as discussed above). The intervention additively applies the semantic axis vector to token positions corresponding to the target word, which appears twice in the prompt, once in the USER prompt and once in the ASSISTANT stem. This conservative test ensures that steering does not merely force an unlikely next-token to become more likely by intervening on its generation. Instead, the steering only influences the next token (the final token in the sequence) indirectly by changing the feature representation of the word being rated. An example of this process is represented in Figure \ref{fig:prompt_schematic}.

The hidden state $\mathbf{w}$ is modified as $\mathbf{w}' = \mathbf{w} + \alpha \lVert \mathbf{w} \rVert \hat{\mathbf{f}}$, where $\hat{\mathbf{f}}$ is the unit-normalized steering vector for the intervention axis and $\alpha$ is a coefficient specifying the intervention's magnitude as a fraction of the local residual stream norm. This norm-relative formulation ensures that each axis receives a proportionally equal intervention regardless of the raw steering vector's magnitude. We set $\alpha = 0.33$ for all interventions, based on trials that showed that this level of intervention influenced outputs without them devolving into incoherence. For each prompt, we intervene on each of the 31 "off-target" semantic axes in both directions (adding or subtracting $\hat{\mathbf{f}}$ to steer once toward \textit{soft} and once toward \textit{hard}, for example).

The spillover effect for a given trial is the difference between $p(pos\_antonym)$ for the steered versus baseline prompt. Differences are aggregated at the axis-pair level. Interventions in the negative direction of the semantic axis are multiplied by -1 to aggregate their effects with the positive direction. Finally, to test if semantic axes that are geometrically aligned exhibit greater spillovers, we calculate the correlation between spillover effects and the cosine similarity between the target and off-target semantic axes.

\section{Results}

\begin{figure}[h!]
    \centering
    \includegraphics[width=\textwidth]{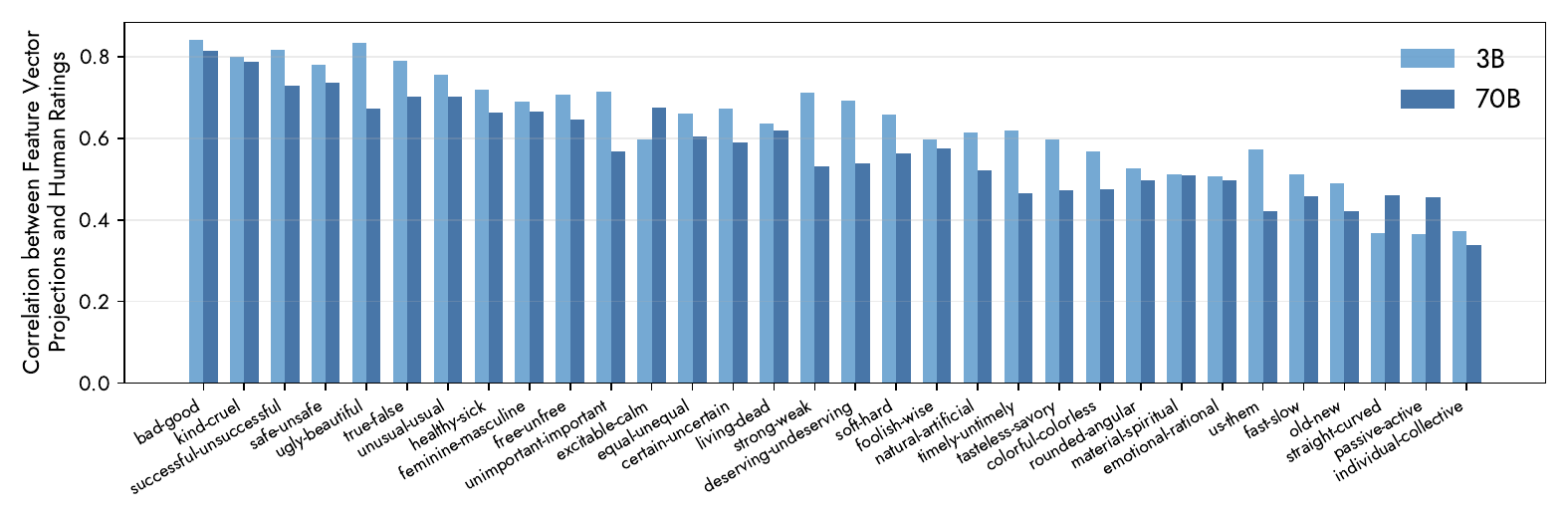}
    \caption{Pearson correlations between human ratings and feature vector projections for 360 words on 32 semantic axes; Llama 3.2 3B and Llama 3.1 70B.}
    \label{fig:llama_proj_corr}
\end{figure}

We begin by examining the correlation between human ratings on the 32 semantic scales and the projections of the corresponding feature vectors on the respective semantic axes. We display associations for small and large models from the Llama 3 family in Figure~\ref{fig:llama_proj_corr}. Across the board, we see substantial positive correlations, with the strongest exhibiting correlations of $r > 0.8$ and the weakest still having an association of $r > 0.3$. Considering the very different methods for measuring these semantic associations, their correlations are notably strong. Association strength is similar between the small and large models, albeit slightly greater in the small model.

\begin{figure}[t]
    \centering
    \includegraphics[height=0.35\textheight]{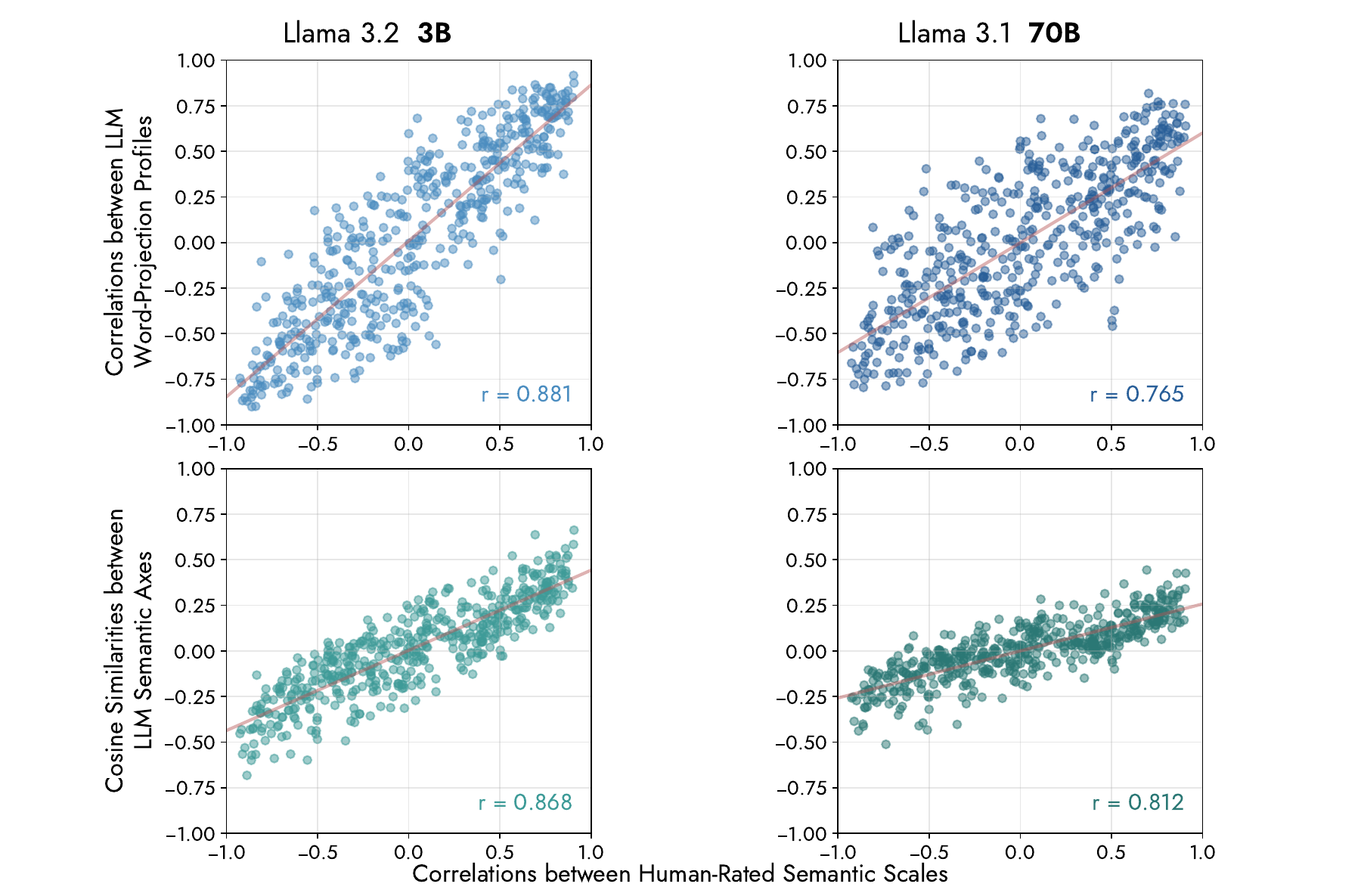}
    \caption{Scatter plots representing pairs of semantic axes by their Pearson correlation in the survey data (x-axis) plotted against the correlations between word feature projections on the respective semantic axes (top) and the cosine similarity between the semantic axes in LLM feature space (bottom); Llama 3.2 3B and Llama 3.1 70B.}
    \label{fig:llama_full_pairwise}
\end{figure}

We next assess whether the correlational structure of the semantic survey is encoded in the geometry of feature space. We calculate the psychological association between semantic scales as the correlation between these scales in the survey data. This produces a pairwise association for each scale. We create pairwise associations between the LLM's semantic axes through two techniques. In the first approach, we take the projections of the 360 word feature vectors onto the 32 semantic axes and calculate the correlations between the semantic axes based on these projections. As a second measure of pairwise association between semantic axes, we calculate cosine similarities between the axes themselves to assess if they are meaningfully non-orthogonal. Figure~\ref{fig:llama_full_pairwise} shows that both of these ways of measuring pairwise similarity between semantic axes closely reproduce the semantic correlational structure derived from the survey.

\begin{figure}[h!]
    \centering
    \includegraphics[width=\textwidth]{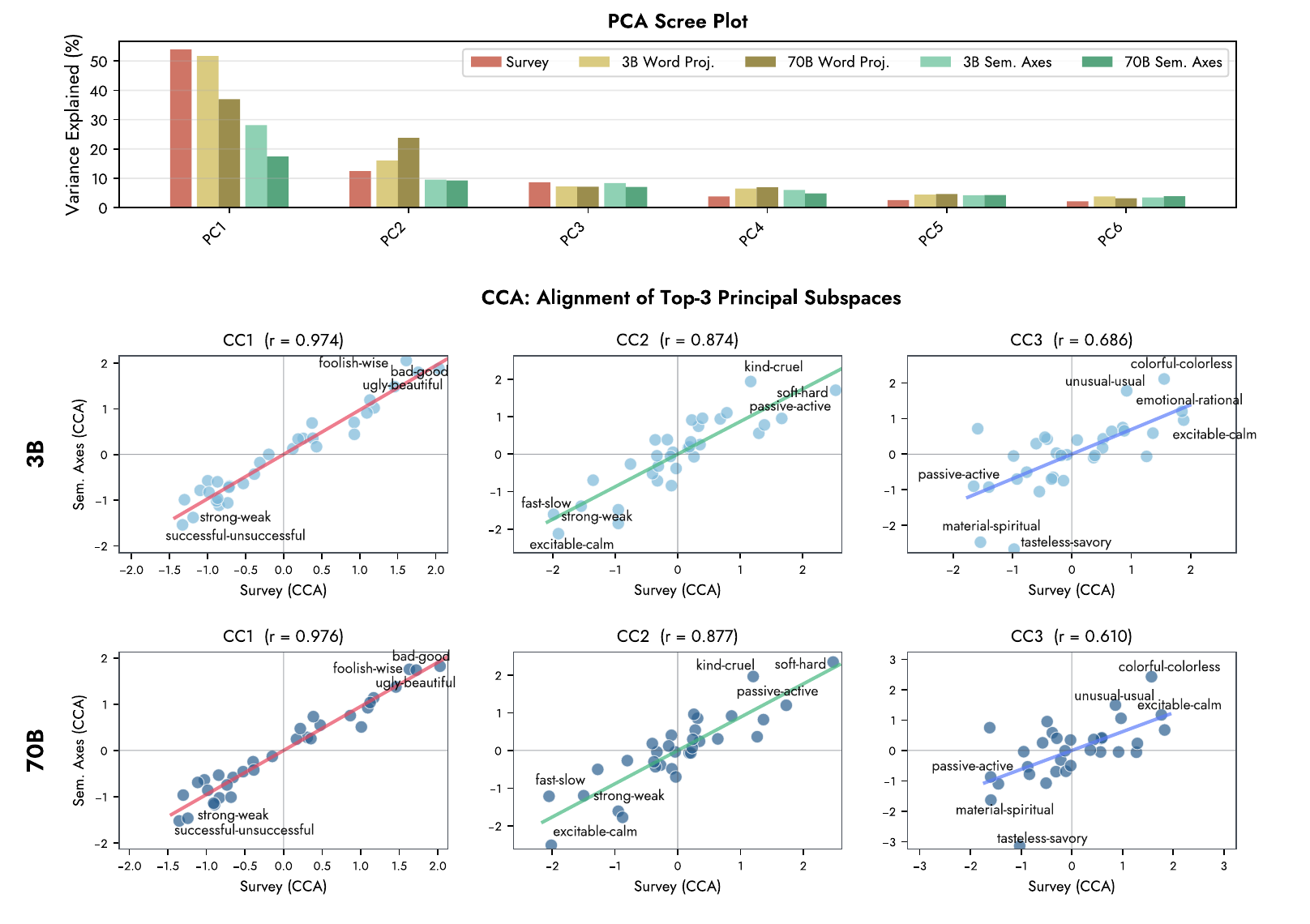}
    \caption{Top: Scree plot from PCA applied to (a) human semantic ratings of 360 words across 32 scales, (b) projection of 360 words feature vectors on semantic axes, and (c) raw semantic axis vectors. Bottom: Canonical Correlation Analysis (CCA) applied to the top-3 PCA subspaces of human survey ratings and LLM semantic axis vectors. Each point represents one semantic axis, plotted by its loading on the corresponding canonical variate in the survey (x-axis) and LLM (y-axis) spaces. High canonical correlations indicate alignment of dimensions of the rotated subspaces; Llama 3.2 3B and Llama 3.1 70B.}
    \label{fig:llama_pca}
\end{figure}

Given that semantic features in LLMs exhibit substantial geometric alignment, we next investigate the extent to which the 32 semantic axes can be adequately represented in a low-dimensional subspace. Figure~\ref{fig:llama_pca} presents results from principal components analysis (PCA) applied to both the survey data and the LLM semantic features. The survey follows the pattern described in prior psychological research – the first three components explain a majority of the variation across the 32 scales measured. The LLM features also show that a large share of variance is explained in the first components. The projections of the word feature vectors on the 32 semantic axes similarly show more than 70\% of the total variance explained by the first three components. But as a more difficult test, we also apply PCA to the raw dimensions of the semantic axes ($d = 3{,}072$ for the small model and $d =8{,}192$ in the large model). We find that considerable variance is explained in the first several components derived by this approach as well. It is important to note that our PCA is constrained because there are only 32 columns in our data matrix. Still, if the semantic axes were orthogonal, there would be no shared variance between them and each would explain 1/32 = 3.1\% of the total variance. Instead, the first three components jointly explain $>45\%$ of the variance in the 3B model and $>33\%$ in the 70B model. 

We next examine how semantic features project onto the most explanatory dimensions in each space. Initial tests showed that the individual principal components produced by PCA were not consistently aligned between the survey and LLM data, but the three components jointly define a similar subspace. To assess this, we use canonical correlation analysis (CCA), which finds linear projections of the two subspaces that are maximally correlated. We find high canonical correlations across all three dimensions, indicating that the subspace spanned by the first three principal components of the LLM’s semantic geometry closely mirrors the correlational structure of human semantic judgments. Comparisons using unrotated principal components (see Appendix) show that while the first component corresponds closely to positive–negative evaluation in both datasets, the second and third components are less well aligned.

\begin{figure}[h!]
    \centering
    \includegraphics[width=\textwidth]{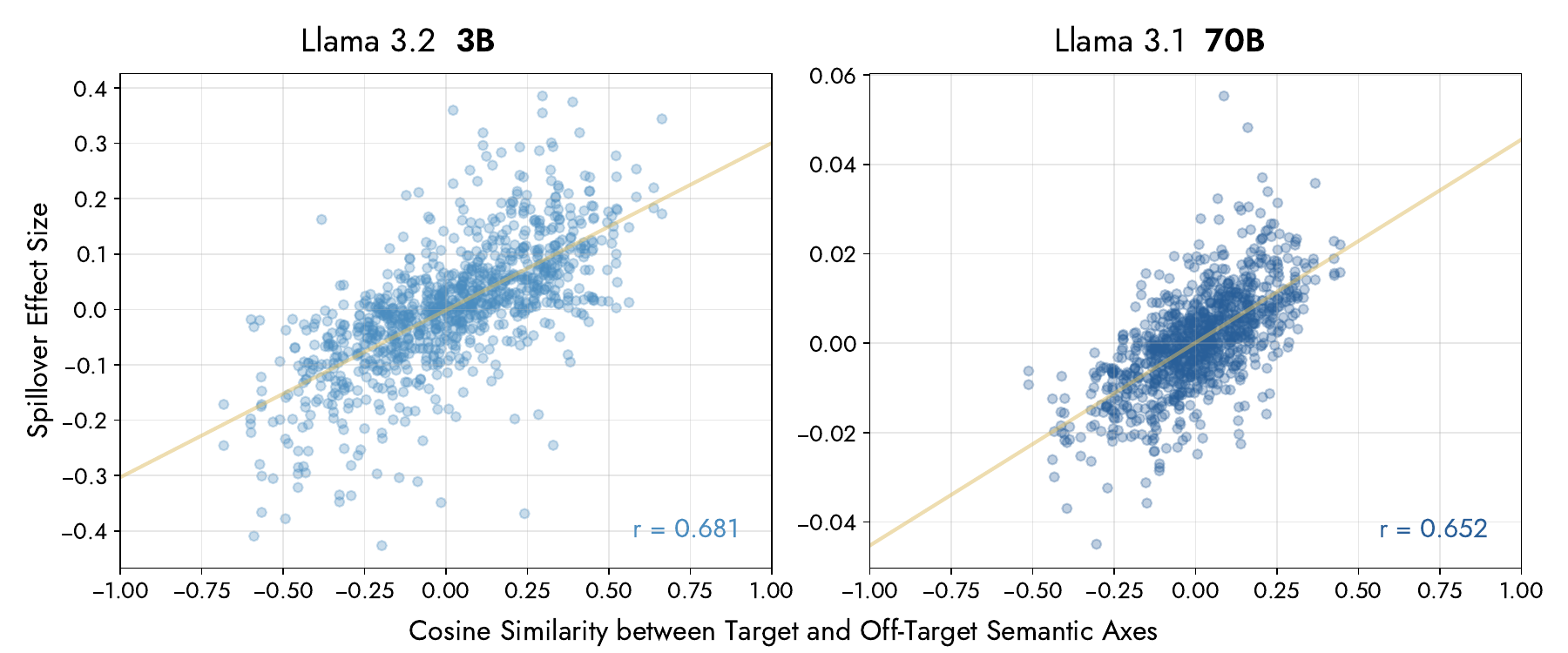}
    \caption{Scatter plots representing pairs of semantic axes by their cosine similarity (x-axis) and the average spillover effect size between them (y-axis); Llama 3.2 3B and Llama 3.1 70B.}
    \label{fig:llama_spillover}
\end{figure}

Next, we explore the implications of the geometry of feature space for model behavior by intervening on semantic axes and assessing off-target spillover effects. Results in Figure~\ref{fig:llama_spillover} show that the geometric alignment of semantic axes is strongly predictive of the direction and magnitude of spillover effects; if one semantic axis has a high cosine similarity with another, an intervention on the target axis is likely to have a spillover effect on the off-target axis. These results suggest that the geometric alignment between feature vectors is substantial enough to influence model behavior, especially in contexts of steering. We note that the magnitude of off-target spillover effects is much smaller for the 70B model than for 3B, but additional analyses included in the Appendix show that even "on-target" steering effects are similarly weaker in the large model than in the small. This suggests that the weaker off-target effects in the large model reflect the model's resilience against steering more than the independence of its semantic features. 

\section{Discussion and Related Work}

To this point, linear features have largely been examined in isolation for understanding and modifying attributes of interest like deception or refusal \citep{arditi2024refusal,templeton2024scaling}. However, classic studies of word embedding models famously show that distributed representations encode meaningful associations between words as proximities and shared directions in latent space \citep{mikolov2013distributed}. Moreover, psychological research suggests that the features humans use to understand and communicate are themselves highly intercorrelated, further suggesting that key features in semantic space are unlikely to be rendered orthogonally \citep{osgood1957measurement}.

Indeed, our findings suggest that LLM internal representations geometrically encode human semantic structure. This finding is directly relevant for steering; we show that steering along one feature is likely to have predictable off-target effects on other features to the extent that they are geometrically aligned. Our findings also have broader relevance for AI safety research. If many safety-relevant features like ``true-false,'' ``good-bad,'' and ``safe-dangerous'' lie on a low dimensional subspace, this could simplify the task of controlling and auditing them. Yet more generally, this geometric alignment could also have implications for how desirable (or undesirable) behaviors generalize across features \citep{betley2025emergent}. Moreover, our findings push against the tendency to consider features as a catalogue of independent and isolated factors \citep{lieberum2024gemma}. Improving our understanding of how LLMs conceptualize good versus bad, wise versus foolish, or beautiful versus ugly may require that we investigate how these features are entangled in the model’s internal representations. Mapping the geometric structure of linear semantic features makes a first step in this direction.

\subsection{Linear Features and Representation Engineering in LLMs}

A central finding in recent interpretability research is that many high-level properties of large language models are approximately linearly represented in activation space. This perspective has been developed most explicitly in work on the linear representation hypothesis, which connects linear probes, contrastive directions, and steering interventions within a common geometric framework \citep{park2023linear}. Follow-up work has taken some strides toward describing geometric relations between features other than orthogonality \citep{park2024geometry}. This work explores how mutually-exclusive categories form polytope structures, but does not consider how related linear features may exhibit meaningful geometric alignment. Other studies suggest that complex features like refusal or truthfulness form multidimensional representations in the activation space \citep{wollschlager2025geometry,ying2026truthfulness}. While this work extends in the direction of multidimensional feature structures, it ultimately seeks to describe a single latent concept rather than describe structural relations between concepts we conventionally consider to be distinct.

\subsection{Distributed Representations and Semantic Structure}

Our approach also connects to the older literature on distributed representations of language. Studies of classic embedding models like word2vec or GLoVe famously showed distributed representations can recover a vast variety of semantic and syntactic features from unstructured web text data, encoding them as proximities and directions in the model's latent space \citep{mikolov2013distributed,mikolov2013efficient,pennington2014glove}. 

A particularly relevant strand of this literature treats induced semantic axes as socially and culturally meaningful objects rather than universal descriptive attributes. \citet{caliskan2017semantics} showed that word embeddings trained on ordinary corpora reproduce a wide range of humanlike biases, while \citet{garg2018word} used embedding geometry to measure historic changes in gender and ethnic stereotypes. In a related vein, \citet{kozlowski2019geometry} argued that projections onto difference vectors can recover broader "cultural dimensions" such as class, gender, and status, and \citep{boutyline2023school} showed that angles between such "cultural dimensions" in sequentially trained word embeddings capture long-term changes in how the relationship between gender and intelligence was understood.

Despite this long and fruitful strand of research with static embeddings, there remains surprisingly little scholarship applying these methods and insights to the activation spaces of LLMs. Our study shows that such analyses can be productively applied to LLMs' internal representations, opening new modes for examining relations between concepts in LLMs.

\subsection{Superposition, Interference, and Non-Orthogonality}

LLM performance along many metrics scales as a function of model size, in part because latent spaces with more dimensions allow for more features to be represented. While the number of features that can be stored orthogonally is equivalent to the model’s dimensionality at a given layer, the number of ``almost orthogonal'' vectors that can be represented in a space scales exponentially with dimensionality \citep{vershynin2018high}, meaning that a vastly greater number of features can be stored with relatively low interference. Early interpretability work with LLMs argued that models exploit the tremendous availability of almost-orthogonal directions to encode multiple features in ``superposition'' \citep{elhage2022superposition}. More recent work, continuing this line of reasoning, has aimed to quantify the maximum number of features that a layer of an LLM can represent while still permitting their recovery by distinct probes \citep{garg2026many}.

However, classic theorizing of distributed representations recognized that dense representations are effective not only because they open up space for many "almost orthogonal" features, but because they allow for meaningful non-orthogonality between related features. \cite{hinton1986distributed} explicitly note that this non-orthogonality is key to "one of the most interesting properties of distributed representations: They automatically give rise to generalizations" (p.82).

Our study is motivated by the perspective that LLMs do not strive to approximate orthogonality between important features, but instead use geometric alignment to encode relations and share information. Our findings offer evidence of this among a relatively small set of semantic features that are paired with human ratings for validation. More broadly, these findings demonstrate that the relational analyses of features that were common with static embeddings are likely to once again be fruitful with the deep representations of LLMs.

\bibliography{references}
\bibliographystyle{colm2026_conference}

%%%%%%%%%%%%%%%%%%%%%%%%%%%%%%%%%%%%%%%%%%%%%%%%%%%%%%%%%%%%

\newpage
\appendix
\section{Supplemental Materials}

\begin{figure}[h!]
    \centering
    \includegraphics[width=\textwidth]{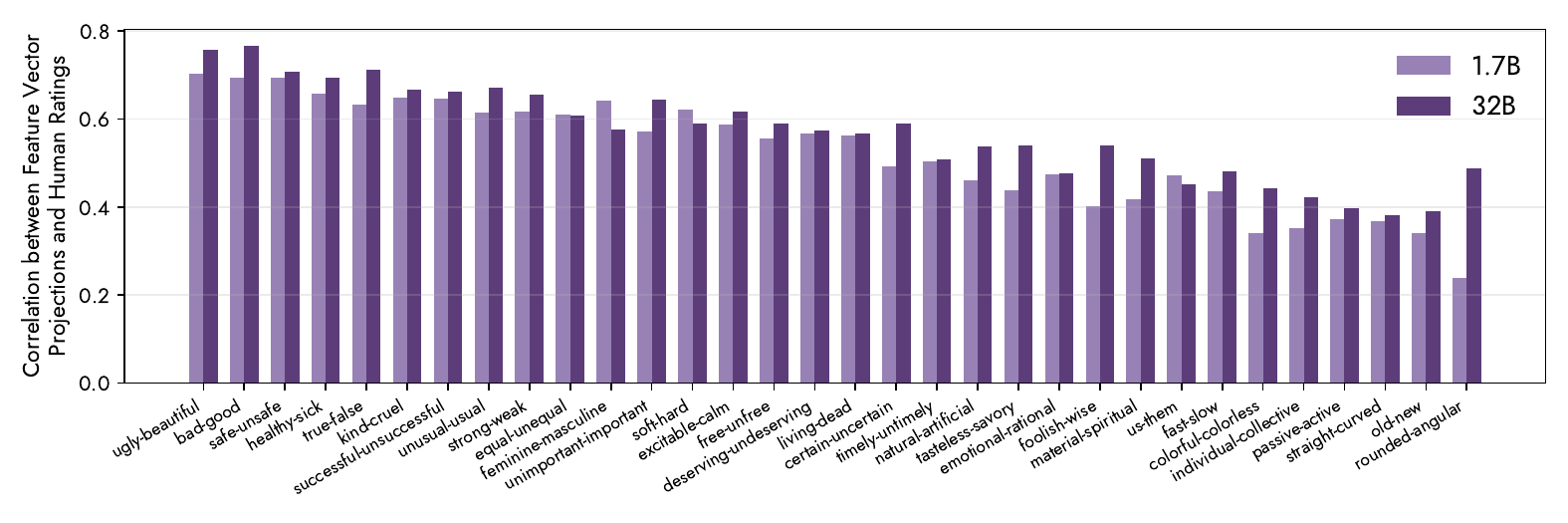}
    \caption{Pearson correlations between human ratings and feature vector projections for 360 words on 32 semantic axes; Qwen3 1.7B and Qwen3 32B.}
    \label{fig:qwen_proj_corr}
\end{figure}

\begin{figure}[h!]
    \centering
    \includegraphics[width=\textwidth]{figures/qwen_full_pairwise.pdf}
    \caption{Scatter plots representing pairs of semantic axes by their Pearson correlation in the survey data (x-axis) plotted against the correlations between word feature projections on the respective semantic axes (top) and the cosine similarity between the semantic axes in LLM feature space (bottom); Qwen3 1.7B and Qwen3 32B.}
    \label{fig:qwen_full_pairwise}
\end{figure}

\begin{figure}[t]
    \centering
    \includegraphics[width=\textwidth]{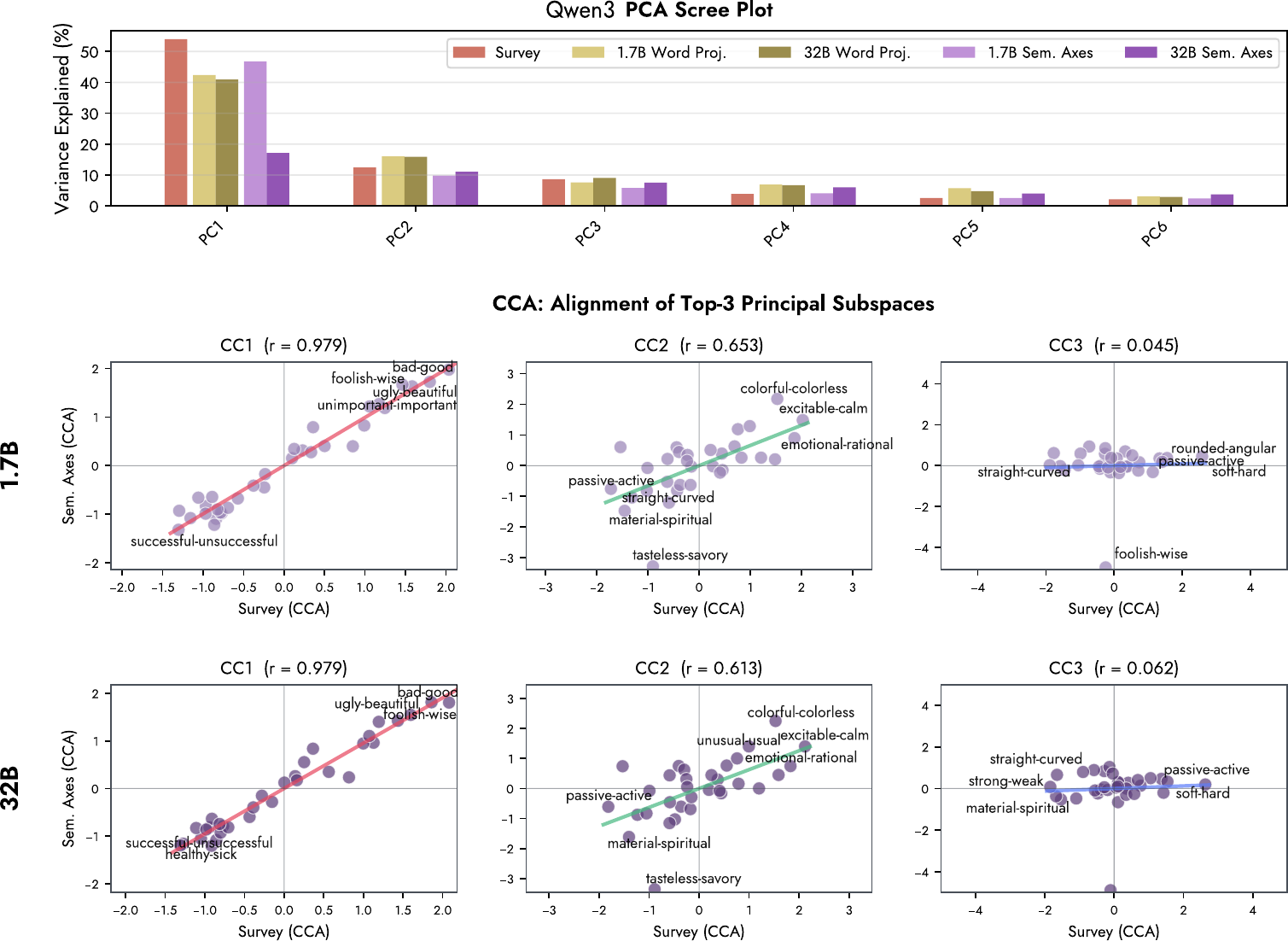}
    \caption{Top: Scree plot from PCA applied to (a) human semantic ratings of 360 words across 32 scales, (b) projection of 360 words feature vectors on semantic axes, and (c) raw semantic axis vectors. Bottom: Canonical Correlation Analysis (CCA) applied to the top-3 PCA subspaces of human survey ratings and LLM semantic axis vectors. Each point represents one semantic axis, plotted by its loading on the corresponding canonical variate in the survey (x-axis) and LLM (y-axis) spaces. High canonical correlations indicate alignment of dimensions of the rotated subspaces; Qwen3 1.7B and Qwen3 32B.}
    \label{fig:qwen_pca}
\end{figure}

\begin{figure}[h!]
    \centering
    \includegraphics[width=\textwidth]{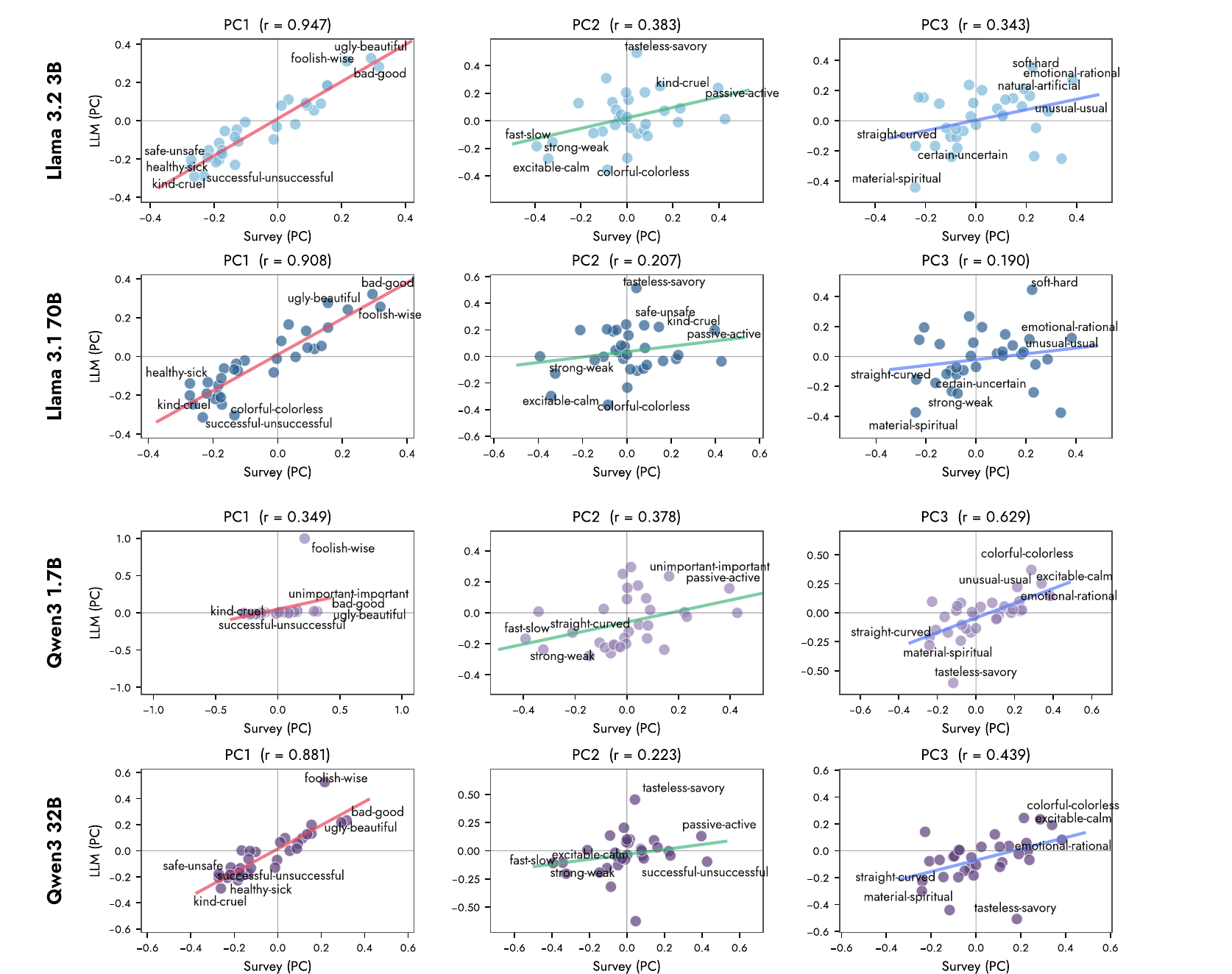}
    \caption{Association between factor loadings from Principal Components Analysis (PCA) applied to the human survey ratings and LLM semantic axis vectors. Each point represents one semantic axis, plotted by its loading on the corresponding principal component in the survey (x-axis) and LLM (y-axis) spaces. Llama 3.2 3B, Llama 3.1 70B; Qwen3 1.7B and Qwen3 32B}
    \label{fig:unrotated_pca}
\end{figure}

\begin{figure}[h!]
    \centering
    \includegraphics[width=\textwidth]{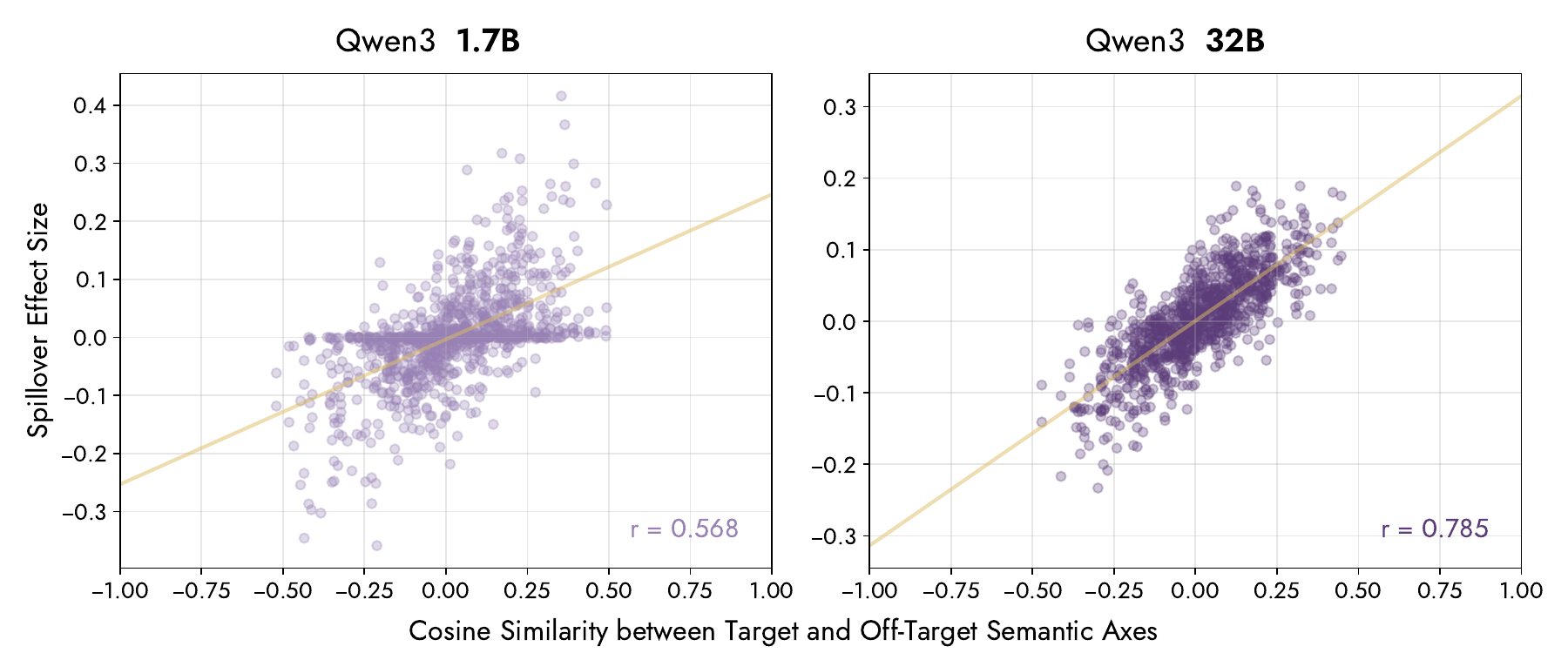}
    \caption{Scatter plots representing pairs of semantic axes by their cosine similarity (x-axis) and the average spillover effect size between them in log-odds (y-axis); Qwen3 1.7B and Qwen3 32B.\\NOTE: See Figure~\ref{fig:qwen_spillover_logit} for explanation of clustering around zero in Qwen3 1.7B}
    \label{fig:qwen_spillover}
\end{figure}

\begin{figure}[t!]
    \centering
    \includegraphics[width=0.85\textwidth]{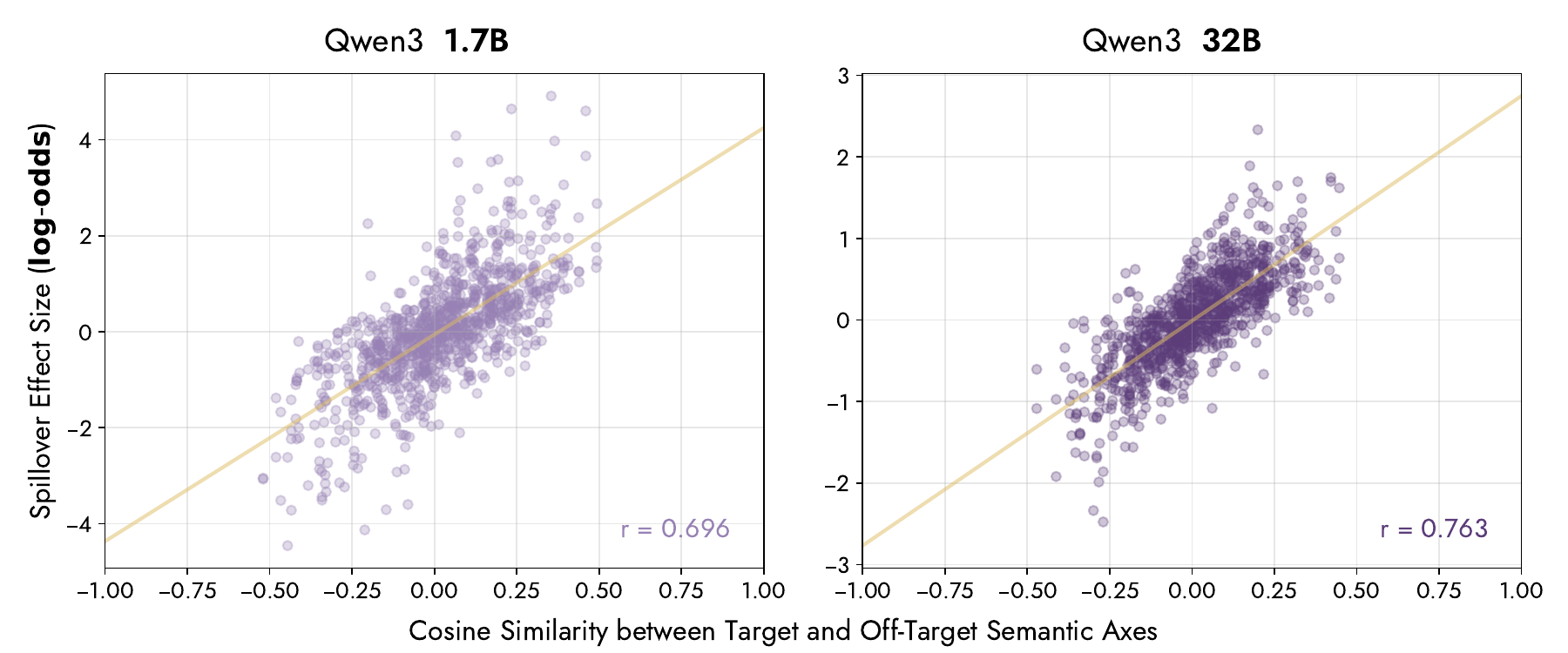}
    \caption{Scatter plots representing pairs of semantic axes by their cosine similarity (x-axis) and the average spillover effect size between them (y-axis); Qwen3 1.7B and Qwen3 32B.\\ NOTE: Results from the spillover analysis on Qwen3 1.7B in Figure~\ref{fig:qwen_spillover} indicate many axis pairs for which the spillover effect is close to zero. However, this is in fact because many of the baseline probabilities are close to 0 or 1. This means a doubling or halving probability can manifest as a change of only 0.01 or 0.02 on the y-axis. This figure shows that, when effects are measured as change in logits rather than probability, the clustering around zero disappears.
}
    \label{fig:qwen_spillover_logit}
\end{figure}

\begin{figure}[t!]
    \centering
    \includegraphics[width=0.85\textwidth]{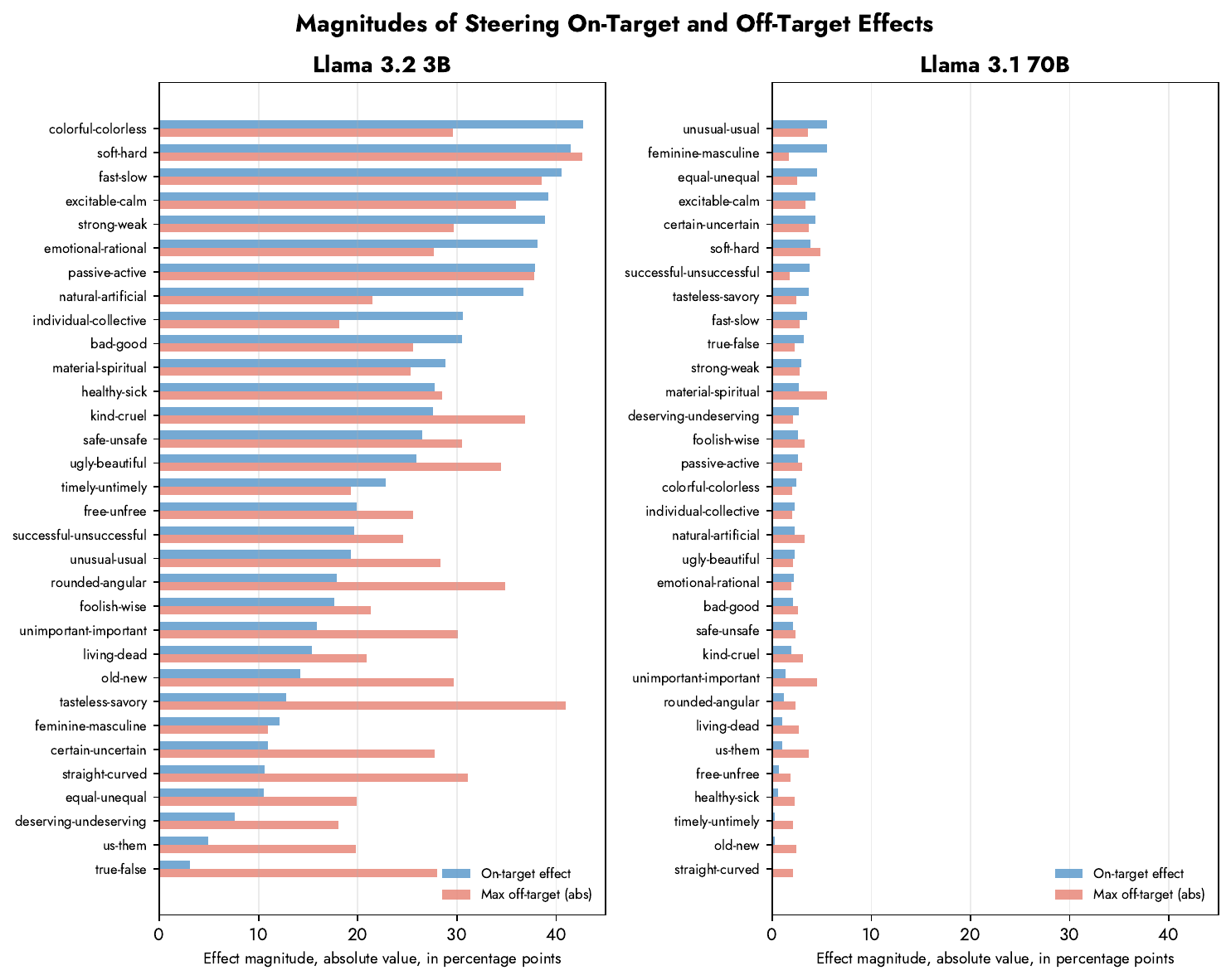}
    \caption{Magnitudes of on-target steering effects (blue) and maximum off-target spillover (red) for each of 32 semantic axes, measured in percentage points. For each axis, the on-target effect is the mean shift in the model's forced-choice probability when intervening on an axis and measuring that same axis; the off-target effect is the largest absolute spillover observed across all other axes. Both panels use the same x-axis scale to highlight the ~10$\times$ difference in steering susceptibility; Llama 3.2 3B and Llama 3.1 70B.}
    \label{fig:spillover_magnitude}
\end{figure}

\end{document}